\documentclass[10pt,twocolumn,letterpaper]{article}

\usepackage{iccv}
\usepackage{times}
\usepackage{epsfig}
\usepackage{graphicx}
\usepackage{amsmath}
\usepackage{amssymb}
\usepackage{caption}
\usepackage{subcaption}
\usepackage{makecell}
\usepackage{interval}
\usepackage{booktabs}
\usepackage{multirow}
\usepackage{textcmds}
\usepackage{textcmds}
\usepackage{stfloats}
\usepackage{float}

\usepackage[pagebackref=true,breaklinks=true,letterpaper=true,colorlinks,bookmarks=false]{hyperref}

\iccvfinalcopy 

\ificcvfinal\fi

\begin{document}

\title{Hybrid-CSR: Coupling Explicit and Implicit Shape Representation \\ for Cortical Surface Reconstruction}

\author{Shanlin Sun, ~Thanh-Tung Le \\
University of California, Irvine\\
{\tt\small \{shanlins, thanhtul\}@uci.edu}
\and
Chenyu You \\
Yale University \\
{\tt\small chenyu.you@yale.edu}
\and
Hao Tang \\
Meta \\
{\tt\small haotang@fb.com}
\and
Kun Han, ~Haoyu Ma, ~Deying Kong, ~Xiangyi Yan, ~Xiaohui Xie \\
University of California, Irvine\\
{\tt\small \{khan7, haoyum3, deyingk, xiangyy4, xhx\}@uci.edu}
}

\maketitle 
\ificcvfinal\thispagestyle{empty}\fi

\begin{abstract}
   We present Hybrid-CSR, a geometric deep-learning model that combines explicit and implicit shape representations for cortical surface reconstruction. 
   Specifically, Hybrid-CSR begins with explicit deformations of template meshes to obtain coarsely reconstructed cortical surfaces, based on which the oriented point clouds are estimated for the subsequent differentiable poisson surface reconstruction. 
   By doing so, our method unifies explicit (oriented point clouds) and implicit (indicator function) cortical surface reconstruction. Compared to explicit representation-based methods, our hybrid approach is more friendly to capture detailed structures, and when compared with implicit representation-based methods, our method can be topology aware because of end-to-end training with a mesh-based deformation module.
   In order to address topology defects, we propose a new topology correction pipeline that relies on optimization-based diffeomorphic surface registration.
   Experimental results on three brain datasets show that our approach surpasses existing implicit and explicit cortical surface reconstruction methods in numeric metrics in terms of accuracy, regularity, and consistency.
   
\end{abstract}

\section{Introduction}

\begin{figure}[!ht]
\centering
\includegraphics[width=0.47\textwidth]{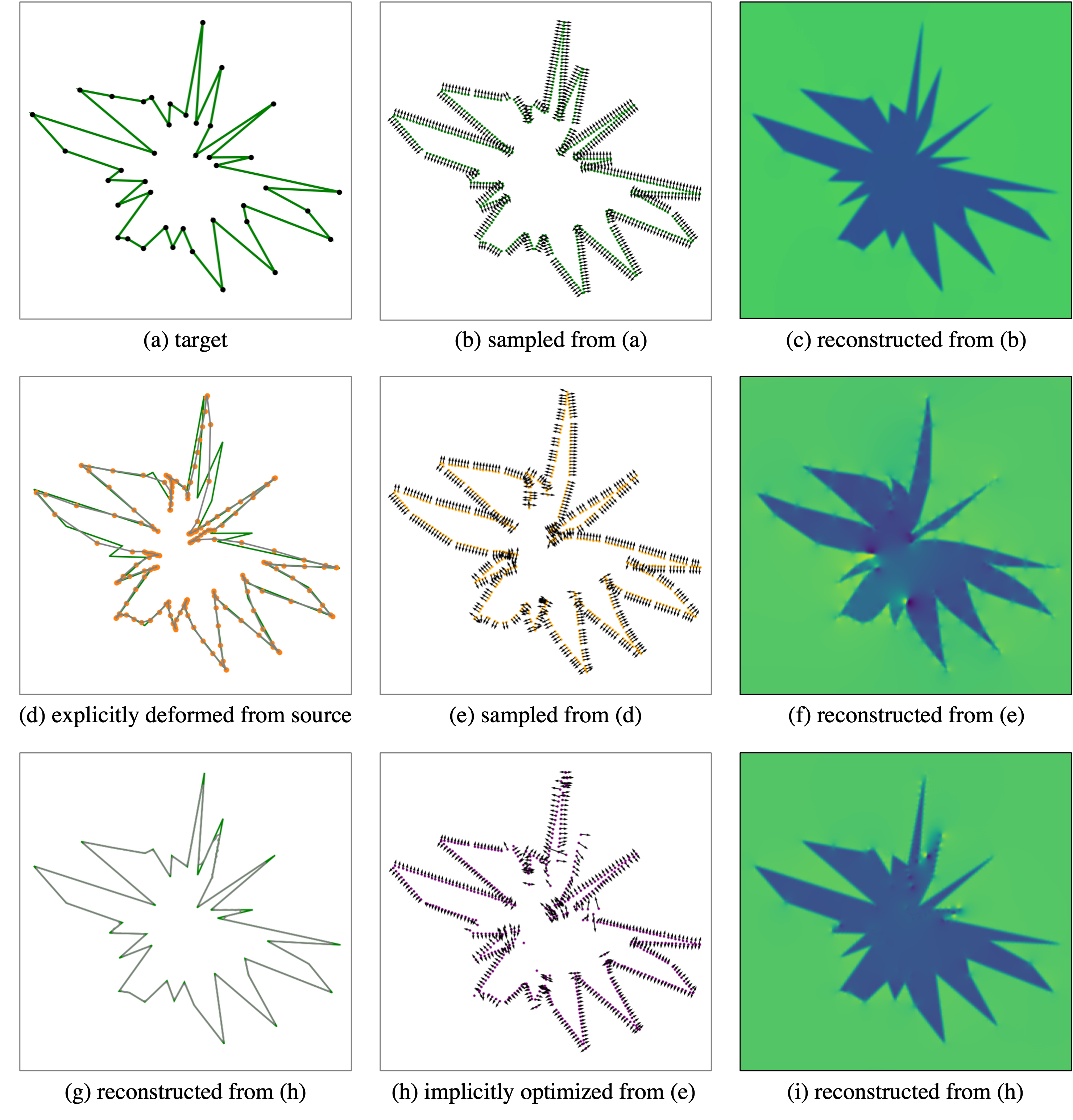}

\caption{\textbf{Contour Representation Toy Example}. 
A polygon target in (a) can be represented by explicitly deforming a circle via optimizing chamfer distance, which gives the orange contour in (d). In contrast, our hybrid method can generate more accurate reconstructed contour in (g) by coupling explicit (oriented point clouds) and implicit shape representation (indicator grid).
} 

\vspace{-1.8em}

\label{fig:toy_example}
\end{figure}

Cortical surface reconstruction is the task to extract both inner and outer surfaces of the cerebral cortex from brain MRI scans, with the inner surface situated between the cortical gray matter (GM) and white matter (WM) while the outer surface between the cerebrospinal fluid and the cortical gray matter.
Accurate and detailed reconstruction of the cortical surface can be used to facilitate brain mapping \cite{destrieux2010automatic, glasser2016multi, toga2003mapping, fischl1999high}, identify biomarkers for neurological disorders\cite{han2006reliability, desikan2010automated, apostolova2010surface}, and enable pre-surgical planning \cite{de2011brain}. However, extracting accurate cortical surface with genus of 0 is still very challenging due to structural complexity of the cerebral cortex and partial volume effect (PVE) \cite{ballester2002estimation} in medical imaging. 

Traditional methods~\cite{dahnke2013cortical, fischl2012freesurfer, fischl1999cortical, kim2005automated} typically first segment the volumetric structures and then extract the cortical meshes via Marching Cubes~\cite{lorensen1987marching,lewiner2003efficient} or level set-based approaches~\cite{osher2004level,li2010distance}.
Specifically, FreeSurfer \cite{fischl2012freesurfer}, the current standard for cortical surface reconstruction, obtains WM surfaces by applying mesh tessellation to segmented WM. 
With the help of convolutional neural network and a novel spherical embedding, FastSurfer \cite{henschel2020fastsurfer} achieves faster and better brain segmentation. While deep learning-based volumetric segmentation models have demonstrated remarkable performance in medical imaging \cite{ronneberger2015u,isensee2018nnu,hatamizadeh2022unetr,cao2023swin,zhou2021nnformer}, the reconstructed cortical meshes may not accurately delineate tissue boundaries due to PVE problem. Furthermore, the predicted brain segmentation may contain topological defects, necessitating time-consuming topology correction algorithms to be applied for genus-0 reconstructed meshes.

One way to overcome the partial volume effect is to utilize deep implicit functions, which can represent complex shapes with fine-grained details by representing the surface implicitly as the zero level-set of continuous functions~\cite{park2019deepsdf, saito2020pifuhd, mescheder2019occupancy, sitzmann2020implicit, tancik2020fourier}. 
Recently, DeepCSR has been proposed to reconstruct the cortical surface leveraging the deep implicit functions in \cite{cruz2021deepcsr}.
As deep implicit functions can be trained efficiently using simple L1 or L2 loss functions, the optimization process is straightforward and robust in terms of convergence.
However, DeepCSR still requires a time-consuming spherical level-set evolution algorithm \cite{pham2010digital} to remove holes and handles from the reconstructed meshes, due to the absence of topology awareness.


Alternatively, there is another branch of work that exploit explicit topology to resolve topology defects and partial volume effects simultaneously~\cite{bongratz2022vox2cortex, lebrat2021corticalflow, santa2022corticalflow++, hoopes2021topofit}.
By progressively deforming explicit meshes from template meshes, explicit cortical reconstruction methods can guarantee reconstructed meshes inherit the desired topology, thus avoiding time-consuming post-processing.
At the same time,  sub-voxel variations of cerebral cortex surfaces can also be captured.
Nonetheless, these methods suffer from two main problems.
Firstly, the often used Chamfer distance loss to align predicted and target meshes tends to get trapped in local minima easily, failing to distinguish bad samples from the true one~\cite{achlioptas2018learning, nguyen2021point}.
Although a weighted Chamfer distance has been proposed to prioritize fitting local regions with high curvatures in Vox2Cortex~\cite{bongratz2022vox2cortex}, the issue is only alleviated but not resolved completely, still resulting in suboptimal assignments between two sets of points \cite{pomerleau2015review, kolouri2018sliced}.
Secondly, while explicit regularizations can reduce self-intersections, they also lead to lower geometric accuracy. 
Neural Mesh Flow~\cite{gupta2020neural} tries to avoid explicit regularization by deforming template meshes through continuous diffeomorphic flow, however, 
it does not perform well in modeling sharp and large deformations. 

In this work, we propose a novel approach for cortical surface reconstruction called Hybrid-CSR, which integrates both implicit and explicit shape representations. 
The method involves an initial step to deform template meshes into a coarsely reconstructed cortical surface, based on which oriented point clouds are estimated for the subsequent implicit cortical surface reconstruction.
Specifically, we apply differentiable Poisson surface reconstruction \cite{Peng2021SAP} to bridge oriented point clouds (explicit) to indicator grids (implicit), from which watertight cortical meshes can be extracted via Marching Cubes. 
In addition, we propose to apply optimization-based diffeomorphic surface registration to realize topology correction.
Our Hybrid-CSR offers several advantages over existing techniques. Unlike voxel-based methods, our approach is not susceptible to partial volume effects. Compared to deep implicit function-based method, Hybrid-CSR produces fewer topology defects and demonstrates superior efficiency in inference. Finally, compared to mesh-based methods, our hybrid approach offers a higher level of expressiveness and flexibility.

The efficacy of our proposed method can be demonstrated by using a simple contour representation toy example, as shown in Fig.~\ref{fig:toy_example}.
The target contour (Fig.~\ref{fig:toy_example}(a)) is a polygon that can be either reconstructed through the deformation of a source contour (circle) or the extraction of the zero level set from an indicator field. 
As can be seen in Fig.~\ref{fig:toy_example}(d), the results obtained from explicit contour deformations optimization are not optimal due to the ``regularizer's dilemma" \cite{gupta2020neural}. 
Instead, our proposed hybrid method optimizes the positions and normals of the oriented points (initialized by Fig.~\ref{fig:toy_example}(e)) and minimizes the difference between the indicator map reconstructed from the ground truth and the optimized oriented point clouds (Fig.~\ref{fig:toy_example}(c)(i)). Hybrid shape representation can produce a well-captured contour from the predicted indicator field, as demonstrated in Fig.~\ref{fig:toy_example}(g). More details about this toy example can be found in the supplementary.

In summary, our main contributions are:

\begin{itemize}
    \item We propose Hybrid-CSR, the first cortical surface reconstruction framework coupling the explicit and implicit shape representation based on differentiable Poisson surface reconstruction.
    
    \item We propose a new topology correction pipeline based on optimization-based diffeomorphic surface registration.
    
    \item We demonstrate on multiple brain datasets that Hybrid-CSR surpasses implicit and explicit reconstruction methods in terms of accuracy, regularity as well as consistency.
\end{itemize}

\section{Related Work}
Traditional cortical surface reconstruction was accomplished through a sequence of image-processing steps, with FreeSurfer \cite{fischl2012freesurfer}, being a widely used approach. While accurate, these methods are constrained by time limitations, with each case taking up to 7-8 hours to complete, thus hindering practical application in clinical settings. Deep learning methods, therefore, are proposed to address the time limitation of traditional approaches and show potential improvement for cortical surface reconstruction task.

\vspace{-0.8em}
\paragraph{Implicit representation} 
FastSurfer\cite{henschel2020fastsurfer} utilizes eigenfunctions of the Laplace-Beltrami operator to parametrize the surface and generate the final spherical map by scaling the 3D spectral embedding vector to unit length.
Furthermore, recent research in 3D computer vision has focused on deep implicit representations \cite{mescheder2019occupancy, sitzmann2020implicit, shipp2007structure, park2019deepsdf}, which have shown great potential in improving the accuracy and efficiency of surface reconstruction. For instance, SegRecon \cite{gopinath2021segrecon} employs a 3D CNN for simultaneously learning segmentation and surface reconstruction by utilizing 3D signed distance function. Additionally, DeepCSR\cite{cruz2021deepcsr} and CortexODE \cite{ma2022cortexode} leverage deep implicit functions to represent the surface, which is then extracted using Marching Cubes \cite{lorensen1987marching, lewiner2003efficient} to produce the output mesh.

\vspace{-0.8em}
\paragraph{Explicit representation} 
Explicit representation methods provide an alternative approach to cortical surface reconstruction, which often learns deformation networks to directly transform a source mesh into a target mesh. For example, CorticalFlow \cite{lebrat2022corticalflow} implements a flow Ordinary Differential Equation (ODE) framework to learn to deform a reference template towards a targeted object. Other methods, such as Voxel2Mesh \cite{wickramasinghe2020voxel2mesh} and Vox2Cortex \cite{bongratz2022vox2cortex}, learn deformable mesh models that take as input a template mesh or sphere initialization and iteratively deform the mesh by learning deformation field of the vertices. 

\vspace{-0.8em}
\paragraph{Diffeomorphic Transformation}
A diffeomorphism is an invertible mapping where the forward and backward transformations are smooth. It is widely used in medical registration problems. Many works \cite{dalca2018unsupervised, balakrishnan2019voxelmorph, krebs2019learning, dalca2019unsupervised, avants2008symmetric} usually assume the velocity field is stationary and defined in the grid space \cite{ashburner2007fast} so that they can apply scaling and squaring method \cite{arsigny2006log} to do fast integration. CorticalFlow and CorticalFlow++ \cite{santa2022corticalflow++} learn a discrete stationary velocity field and integrate the deformations associated with the template meshes using interpolation and traditional ODE solvers.
Recently, with the power of neural ordinary differential equation solver \cite{chen2018neural, chen2020learning}, optimizing a neural diffeomorphic flow efficiently became possible. NDF \cite{sun2022topology} addresses organ shape representation and registration simultaneously by decomposing the implicit shape representation into continuous diffeomorphic transformations and template shape representation. Neural Mesh Flow\cite{gupta2020neural} focuses on generating manifold mesh from images or point clouds via conditional continuous diffeomorphic flow. 
\section{Method}

\begin{figure*}
\centering
\includegraphics[width=\textwidth]{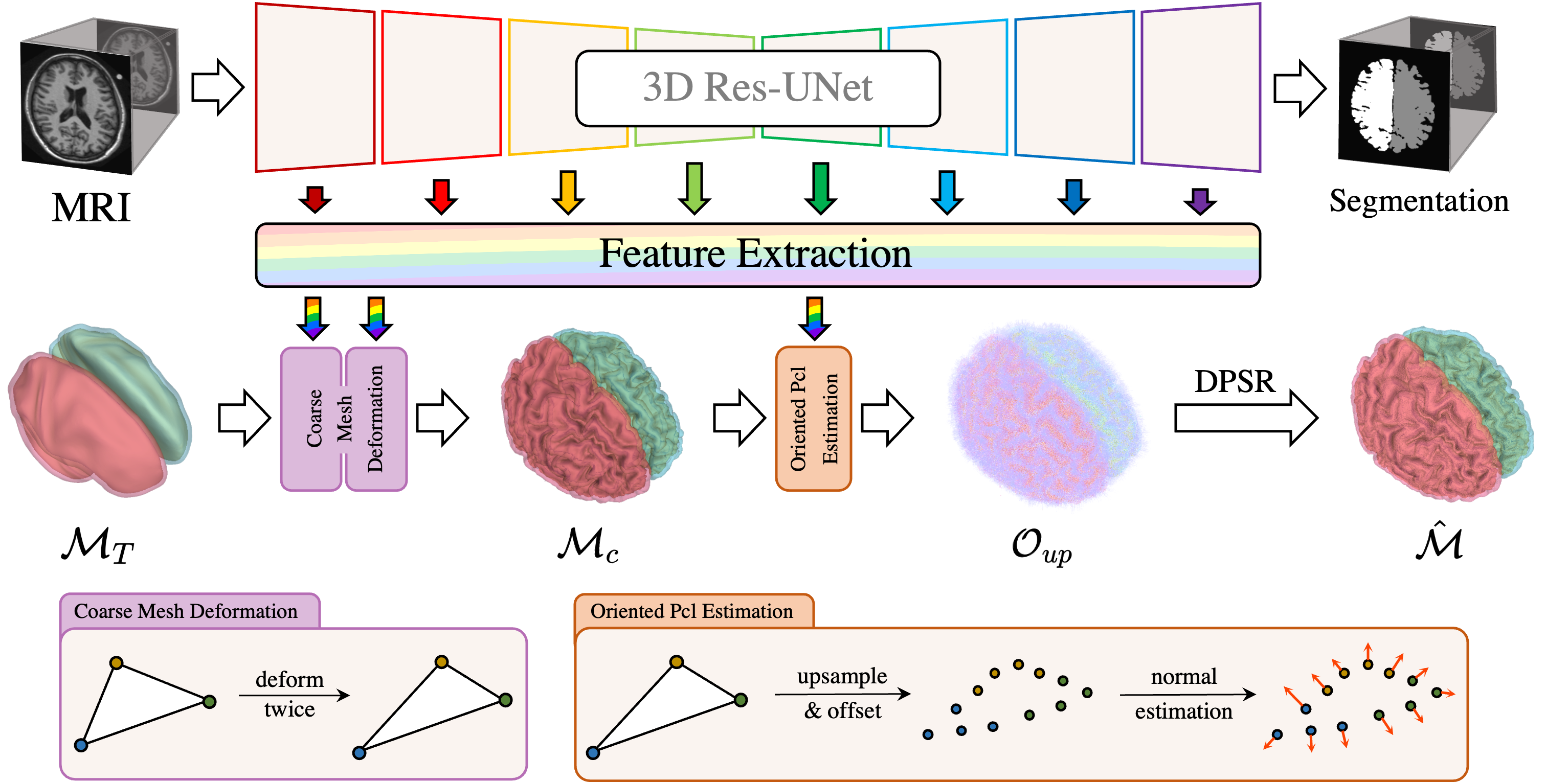}

\caption{\textbf{The overview of Hybrid-CSR}. The architecture takes as input a 3D MRI and template meshes and predicts a voxel-wise segmentation map and cortical surface meshes. This illustration presents the first two steps of our whole pipeline: coarse mesh deformation and oriented point cloud estimation. The coarse mesh deformation module offsets the vertices of template meshes twice using GCN based on image and shape descriptive features. The oriented point cloud estimation module predicts the positions and normals sequentially with GCN-based feature encoder and GLU-based final layer.}

\label{fig:framework}
\end{figure*}

In this section, we will provide a comprehensive description of our proposed method, Hybrid-CSR, which aims to couple explicit and implicit surface representations to enable accurate cortical surface reconstruction from 3D MRI. The pipeline of Hybrid-CSR consists of three main parts, including coarse mesh deformation (Sec.~\ref{sec:coarse_deform}), oriented point cloud estimation (Sec.~\ref{sec:opcl_est}), topology correction (Sec.~\ref{sec:tc}) and surface refinement (Sec.~\ref{sec:refine}). 
Prior to delving into the specifics of our method, we will first describe the methods of reconstructing surfaces from oriented point clouds using differentiable Poisson surface reconstruction (Sec.~\ref{sec:dpsr}) and modelling continuous deformations using diffeomorphic flow (Sec.~\ref{sec:cdf}).

\subsection{Preliminary Knowledge}
\label{sec:background}

\subsubsection{Poisson Surface Reconstruction}
\label{sec:dpsr}
 Poisson surface reconstruction (PSR) \cite{kazhdan2006poisson} aims to recover an indicator function $\chi \in \mathbb{R}^n$ from K sampled points $\mathcal{P}=\left\{p_i \in \mathbb{R}^3\right\}_{i=1}^K$ with normals $\mathcal{N}=\left\{n_i \in \mathbb{R}^3\right\}_{i=1}^K$, by satisfying that $\chi$ changes sharply between positive and negative values at the surface boundary along the direction orthogonal to the surface. We consider the case where $n := r \times r \times r$, and $d=3$,  where $r$ is the resolution of the indicator grid. In practice, PSR first constructs a point normal field $\mathbf{q} \in \mathbb{R}^{n \times d}$ from $\mathcal{P}$ and $\mathcal{N}$.  Then, it formulates the reconstruction of $\chi$ as a Poisson equation: $\nabla^2 \chi:=\nabla \cdot \nabla \chi=\nabla \cdot \mathbf{q}$,
 where the Laplacian of $\chi$ is equal to the divergence of the normal vector field $\mathbf{v}$, subject to the boundary condition that $\chi$ is zero at infinity. This is equivalent to minimizing a quadratic energy function such that $\min _\chi\|\nabla \chi-\mathbf{q}\|_2^2$. 

 Unlike \cite{kazhdan2006poisson}, which encodes the indicator function $\chi$ as a linear combination of sparse basis functions and solves the partial differential equation (PDE) using a finite element solver on an octree, Differentiable Poisson Surface Reconstruction (DPSR) \cite{Peng2021SAP} represents $\chi$ on a 3D grid in a discrete Fourier basis and employs a spectral solver \cite{canuto2007spectral}. The unnormalized indicator function $\chi^{\prime}$ is given by
 \begin{equation}
     \tilde{\chi}=\tilde{g}_{\sigma, r}(\mathbf{u}) \odot \frac{i \mathbf{u} \cdot \tilde{\mathbf{q}}}{-2 \pi\|\mathbf{u}\|^2},\quad \chi^{\prime}=\operatorname{IFFT}(\tilde{\chi}) 
    \label{eq:dpsr}
\end{equation}
where spectral domain signal is denoted as tilde symbol, i.e., $\tilde \chi = FFT(\chi)$, $\mathbf{u} \in \mathbb{R}^{n \times d}$ denotes the spectral frequencies, $\tilde{\mathbf{q}}$ represents the fast Fourier transform (FFT) of $\mathbf{q}$, IFFT($\tilde \chi$) represents the inverse FFT of $\tilde \chi$, and $\tilde{g}_{\sigma, r}(\mathbf{u})$ is a Gaussian smoothing kernel of bandwidth $\sigma$ at grid resolution $r$ in the spectral domain. We denote the element-wise product as $\odot : \mathbb{R}^n \times \mathbb{R}^n \mapsto \mathbb{R}^n$, and the dot product $(\cdot): \mathbb{R}^{n \times d} \times \mathbb{R}^{n \times d} \mapsto \mathbb{R}^n$, and the L2-norm as $\|\cdot\|^2 : \mathbb{R}^{n \times d} \mapsto \mathbb{R}^n$ .
Finally, the normalized indicator function is obtained by subtracting the mean of the unnormalized indicator function at $\mathcal{P}_{up}$ and re-scaling it, written as 
 \begin{equation}
    \chi=\frac{m}{\operatorname{abs}\left(\left.\chi^{\prime}\right|_{\mathbf{x}=0}\right)} 
    \left(
    \chi^{\prime}-\frac{1}{|\{\mathcal{P}\}|} 
    \sum_{\mathbf{c} \in\{\mathcal{P}\}} 
    {\chi^{\prime}|_{\mathbf{x}=\mathbf{c}}}
    \right).
\label{eq:dpsr_nor}
\end{equation}

\subsubsection{Diffeomorphic Flow}
\label{sec:cdf}

Diffeomorphic flows can establish dense point correspondences between source and target surfaces, and preserve the desired geometric topology. In our methods, it is used for mesh registration in topology correction pipeline. 
Let $\Phi(\boldsymbol{p}, t): \Omega\subset\mathbb{R}^{3} \times [0, 1] \mapsto \Omega\subset\mathbb{R}^3$ define a continuous, invertible trajectory from the initial position $\boldsymbol{p}=\Phi(\boldsymbol{p}, 0)$ to the final position $\boldsymbol{p}^{\prime}=\Phi(\boldsymbol{p}, 1)$, satisfying such ordinary differential equation (ODE) and the initial condition:
\begin{equation}
    \frac{\partial \Phi(\boldsymbol{p}, t)}{\partial t}=\boldsymbol{v}(\Phi(\boldsymbol{p}, t), t) \quad \text { s.t. } \quad \Phi(\boldsymbol{p}, 0)=\boldsymbol{p},
    \label{eq:forward_ode}
\end{equation}
where $\boldsymbol{v}(\boldsymbol{p}, t): \Omega \times [0, 1] \mapsto \Omega$ indicates the velocity vector of coordinate $\boldsymbol{p}$ at time t. If $\boldsymbol{v}$ is Lipschitz continuous, a solution to Eq.~\ref{eq:forward_ode} exists and is unique in the interval $[0, 1]$, which ensures that any two deformation trajectories do not cross each other \cite{Coddington1984-jh}. 

\subsection{Coarse Mesh Deformation}
\label{sec:coarse_deform}

As is shown in Fig.~\ref{fig:framework}, the coarse mesh deformation module takes template meshes $\mathcal{M}_T=\left(\mathcal{V}_T, \mathcal{E}_T\right)$ as input and deforms them into $\mathcal{M}_c=\left(\mathcal{V}_c, \mathcal{E}_T\right)$ given the image features extracted from the voxel-based neural network. Our coarse mesh deformation module follows the design from Vox2Cortex \cite{bongratz2022vox2cortex}, using the same mesh templates, residual Unet-based segmentation branch \cite{zhang2018road, isensee2018nnu}, GCN-based mesh deformation branch \cite{ravi2020accelerating, bronstein2021geometric}, feature sampling, as well as training losses. However, we deform the template meshes only twice, in contrast to the four steps in Vox2Cortex. Coarse mesh deformation enables a good initialization for the subsequent oriented point cloud estimation, resulting in fewer outlier points. In our experiments, we utilize smaller templates ($\approx 42000$ vertices per surface) VoxCortex provides with their official implementation for both training and inference.

\subsection{Oriented Point Cloud Estimation}
\label{sec:opcl_est}

The oriented point cloud estimation module takes the coarsely deformed meshes $\mathcal{M}_c$ as input and predicts the coordinates and normals of an upsampled oriented point cloud $\mathcal{O}_{up}$ for Poisson surface reconstruction. This is achieved through encoding features on the graph (mesh), followed by the estimation of point positions and normals in sequence.

\vspace{-0.8em}
\paragraph{Feature Encoding} Same as Vox2Cortex, we applied a GCN-based residual block to encode point features based on the image features. The residual block is composed of three GCNs with subsequent batch normalization layers and ReLU activations. The input residuum is added before the last ReLU output and reshaped with nearest-neighbor interpolation if needed. Let $\mathbf{f}_i \in \mathbb{R}^{d_{out}}$ define the encoded feature of a vertex $\boldsymbol{v_i} \in \mathcal{V}_c$.  

\vspace{-0.8em}
\paragraph{Position Estimation} Our method predicts multiple offsets associated with each vertex to upsample the point cloud and improve shape representation accuracy. Instead of using GCNs, which tend to generate over-smoothed offsets, we employ the gated linear unit (GLU) \cite{dauphin2017language} to estimate the displacements $\boldsymbol{d_i} \in \mathbb{R}^{(S \times 3)}$ with an upsample scale of $S$ for each vertex. This can be expressed as:
\begin{equation}
    \boldsymbol{d_i} = (\mathbf{W}_0 \mathbf{f}_i+\mathbf{b}_0) \odot \sigma(\mathbf{W}_1 \mathbf{f}_i+\mathbf{b}_1)
    \label{eq:glu_point}
\end{equation}
where $\mathbf{W}_0, \mathbf{W}_1 \in \mathbb{R}^{(S \times 3) \times d_{out}}$ together with $\mathbf{b}_0, \mathbf{b}_1 \in \mathbb{R}^{(S \times 3)}$ represent linear projections and $\sigma$ represents sigmoid function. The upsampled displacements are then added to the vertex $\boldsymbol{v}_i$ to obtain the point cloud position $\boldsymbol{p}_i^{up} \in \mathbb{R}^{S \times 3}$. In our experiments, $S$ is set to be 7.

\vspace{-0.8em}
\paragraph{Normal Estimation} We begin by extracting multi-scale image features for each point in $\mathcal{P}_{up}$. Features from points that have been displaced from the same vertex in $\mathcal{M}_c$ are stacked together, and point features $\mathbf{f}_i^{up}$ are learned through a residual GCN layer. 
Based on $\mathbf{f}_i^{up}$ and $\boldsymbol{p}_i^{up}$, the normals $\boldsymbol{n}_i^{up} \in \mathbb{R}^{S \times 3}$ are predicted by a GLU, formulated as
\begin{equation}
\begin{split}
    \boldsymbol{n}_i^{up} = (\mathbf{W}_0^{up} [\mathbf{f}_i^{up}, \boldsymbol{p}_i^{up}]+\mathbf{b}_0^{up}) \odot \\
    \sigma (\mathbf{W}_1^{up}[\mathbf{f}_i^{up}, \boldsymbol{p}_i^{up}]+\mathbf{b}_1^{up})
    \label{eq:glu_normal}
\end{split}
\end{equation}
where $\mathbf{W}_0^{up}, \mathbf{W}_1^{up} \in \mathbb{R}^{(S \times 3) \times (d_{out} + S \times 3)}$ together with $\mathbf{b}_0, \mathbf{b}_1 \in \mathbb{R}^{(S \times 3)}$ represent linear projections and $\sigma$ represents sigmoid function. Now, the upsampled oriented point cloud $\mathcal{O}_{up} = (\mathcal{P}_{up}=\{\boldsymbol{p}_i^{up}\}_{i=1}^K, ~ \mathcal{N}_{up}=\{\boldsymbol{n}_i^{up}\}_{i=1}^K)$ is achieved.

\vspace{-0.8em}
\paragraph{Surface Reconstruction}
In practice, the Differentiable Poisson Surface Reconstruction (DPSR) method acquires a uniformly discretized point normal field $\mathbf{v}$ by rasterizing the predicted oriented point cloud normals $\mathcal{N}_{up}$ onto four uniformly sampled 3D grids. Utilizing Eq.~\ref{eq:dpsr_nor}, an $r^3$ indicator grid $\hat{\chi}$ can be obtained for each cortical structure, and subsequently used to reconstruct cortical meshes $\hat{\mathcal{M}}=\left(\hat{\mathcal{V}}, \hat{\mathcal{E}}\right)$ via Marching Cubes.
The smoothness and resolution of the indicator grid $\hat{\chi}$ can be modulated by the values of $\sigma$ and $r$. In our experiments, we set $\sigma$ to 2 and $r$ to 256 for both training and inference.

\vspace{-0.8em}
\paragraph{Training} 
For each cortical surface, we acquire the ground truth indicator grid $\chi$ by running DPSR on densely sampled oriented point clouds sampled from the (pseudo) ground truth meshes. Next, we use a Sobel filter \cite{kroon2009numerical} to detect edges in $\chi$. The resulting edge map is then smoothed using a Gaussian filter with a kernel size of 7 and a standard deviation of 1. Finally, we apply a weighted mean square error (wMSE) to measure the difference between predicted and ground truth indicator grids, especially along the boundary regions, and train Hybrid-CSR end-to-end, given as 
\begin{equation}
    \mathcal{L}_{\text {DPSR}}=\|w_{edge} \odot \left( \hat{\chi}-\chi \right)\|^2
\label{eq:loss}
\end{equation}
where $w_{edge} \in \mathbb{R}^n$ is the smoothed edge map of the ground truth indicator grid. More implementation details are included in the supplementary.

\subsection{Topology correction}
\label{sec:tc}

We propose a pipeline to fix the topological defects present in $\hat{\mathcal{M}}$. The WM meshes are prone to having \qq{holes} while the pial meshes tend to have \qq{handles}, which prevent the reconstructed meshes from being genus-0. Four cortical meshes in $\hat{\mathcal{M}}$ can be post-processed in parallel. For simplicity, we will note $\hat{\mathcal{M}}$ as a single mesh (either WM or pial) in this section.

\begin{figure}
    \centering
    \begin{subfigure}[b]{0.22\textwidth}
        \centering
        \includegraphics[width=\textwidth]{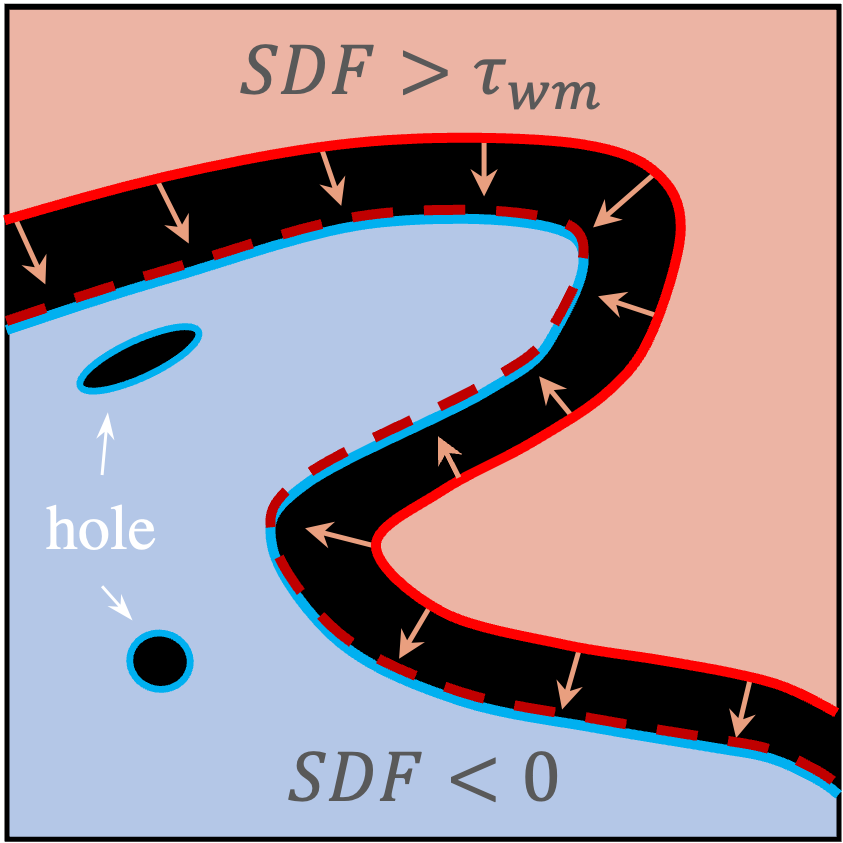}
        \caption{WM}
        \label{fig:wm_topo}
    \end{subfigure}
    \hfill
    \begin{subfigure}[b]{0.22\textwidth}
        \centering
        \includegraphics[width=\textwidth]{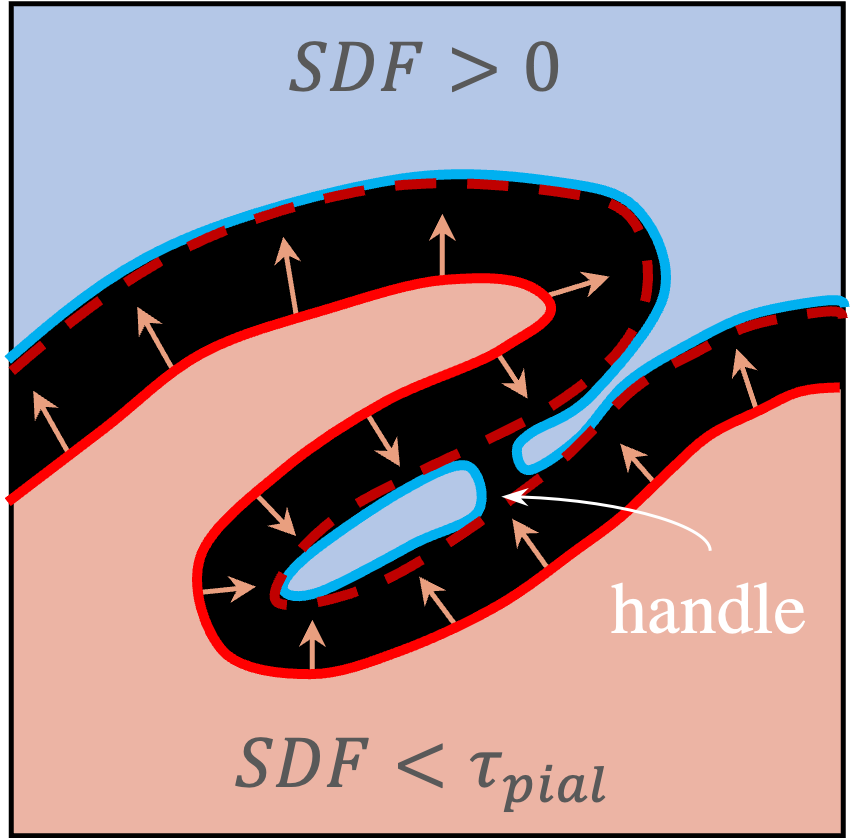}
        \caption{Pial}
        \label{fig:gm_topo}
    \end{subfigure}
    \caption{\textbf{Illustration of Topology Correction and Mesh Registration.} Blue contours represent $\hat{\mathcal{M}}$, red contours represent $\hat{\mathcal{M}}_{tc'}$, and dotted red contour represent $\hat{\mathcal{M}_{tc}}$.}
    \label{fig:topo_corr}
\end{figure}
We first binarize the predicted indicator grids $\hat{\chi}$ into masks, selecting the largest connected component that contains the majority of voxels. The binary masks are then transformed into signed distance grids \cite{butt1998optimum} and smoothed with a Gaussian filter having a standard deviation of 1. 
In Fig.~\ref{fig:topo_corr}, we demonstrate how we transform $\hat{\mathcal{M}}$ into topologically correct $\hat{\mathcal{M}}_{tc'}=(\hat{\mathcal{V}}_{tc'}, \hat{\mathcal{E}}_{tc})$.
To fill the holes in WM meshes (Fig.~\ref{fig:wm_topo}), we extract the $\tau_{wm}$-level set of the smoothed signed distance grid, where $\tau_{wm} = 0.5$. Similarly, to address the handles in pial meshes (Fig.~\ref{fig:gm_topo}), we extract the $\tau_{pial}$-level set of the smoothed singed distance grid, where $\tau_{pial} = -1.8$.

Then, we apply optimization-based diffeomorphic registration to map $\hat{\mathcal{M}}_{tc'}$ to $\hat{\mathcal{M}}$, obtaining the accurate, topologically correct cortical reconstruction results, denoted as $\hat{\mathcal{M}}_{tc}=(\hat{\mathcal{V}}_{tc}, \hat{\mathcal{E}}_{tc})$, as indicated by the dotted red contours in Fig.~\ref{fig:topo_corr}.
To do so, we assume ${\boldsymbol{v}}$ in Eq.~\ref{eq:forward_ode} is stationary and can be modeled via a neural field \cite{xie2022neural, sun2022mirnf}, such that $\mathcal{F}_{\theta}(\boldsymbol{p}) = [\boldsymbol{v}_{p_x}, \boldsymbol{v}_{p_y}, \boldsymbol{v}_{p_z}]^T$, where $\boldsymbol{p}$ and $\theta$ denote the network 3D coordinates and parameters. 
Then, the initial value problem (IVP) in Eq.~\ref{eq:forward_ode} can be solved with a Differentiable ODE Solver (NODE) \cite{chen2018neural} whose dynamic function is set to be $\mathcal{F}_{\theta}$. 

In the forward pass, the destination position $\hat{\boldsymbol{p}}_{tc} \in \hat{\mathcal{V}}_{tc}$ starting from $\hat{\boldsymbol{p}}_{tc'} \in \hat{\mathcal{V}}_{tc'}$ is estimated by integrating $\mathcal{F}_{\theta}(\boldsymbol{p})$ from $t=0$ to $t=1$ via 4th-order Runge–Kutta methods with step size being 0.2. 
For backpropagation, NODE adopts the adjoint sensitivity method \cite{pontryagin1987mathematical}, which retrieves the gradient by solving the adjoint ODE backwards in time and allows solving with O(1) memory usage no matter how many steps the ODE solver takes.
The network parameters $\theta$ are optimized with chamfer distance using the Adam optimizer for 75 iterations with a step size of 3e-4. To compute the chamfer distance, we sample 150,000 points with normals from both $\hat{\mathcal{M}}_{tc}$ and $\hat{\mathcal{M}}_{tc'}$ per iteration. 

\subsection{Learning-based Surface Refinement}
\label{sec:refine}
A learning-based diffeomorphic surface registration model, similar to CortexODE~\cite{ma2022cortexode}, is utilized to generate refined cortical meshes $\hat{\mathcal{M}}_f=(\hat{\mathcal{V}}_f, \hat{\mathcal{E}}_{tc})$ given $\hat{M}_{tc}$. It will remove some artifacts introduced during the topology correction procedures and improve the alignment of the cortical surfaces with the original 3D MRI data.
Different from the CortexODE, our refined pial surfaces $\hat{\mathcal{M}}_f^{pial}$ are transformed from $\hat{\mathcal{M}}_{tc}^{pial}$ instead of $\hat{\mathcal{M}}_{tc}^{wm}$. 

\section{Experiments}
\label{sec:exp}

\subsection{Datasets}
\label{exp:data}
To evaluate the performance of our method on the reconstruction of cortical surfaces from MRI images, we use three publicly available datasets: the Alzheimer's Disease Neuroimaging Initiative (ADNI) dataset \cite{jack2008alzheimer}, the OASIS-1 dataset \cite{marcus2007open}, and the test-retest (TRT) dataset \cite{maclaren2014reliability}. We obtain the pseudo-ground truth surfaces generated from Freesurfer v5.3 \cite{fischl2012freesurfer} for all three datasets. We strictly follow pre-processing pipeline from \cite{bongratz2022vox2cortex}. Specifically, we first register the MRIs to the MNI152 scan. After padding the input images to have shape $192 \times 208 \times 192$, we resize them to $128 \times 144 \times 128$. The intensity values are min-max-normalized to the range $[0,1]$.

\vspace{-0.8em}
\paragraph{ADNI} We use a subset of the ADNI dataset \cite{jack2008alzheimer} containing a total of $419$ T1-weighted (T1w) brain MRI from subjects aged from $55$ to $90$ years old. We stratify the dataset into $299$ scans for training ($\approx 70\%$), $40$ scans for validation($\approx 10\%$), and $80$ scans for testing ($\approx 20\%$). We report all of our experiment results on the test set.

\vspace{-0.8em}
\paragraph{OASIS} For the OASIS dataset \cite{marcus2007open}, we use all of $416$ T1-weighted (T1w) brain MRI images. We stratify the dataset into $292$ scans for training ($\approx 70\%$), $44$ scans for validation ($\approx 10\%$), and $80$ scans for testing ($\approx 20\%$). We report all of our experiment results on the test set.

\vspace{-0.8em}
\paragraph{Test-retest} To analyze the consistency of our approach, we evaluate all $120$ scans from three different subjects, where each subject is scanned twice in $20$ days.

\subsection{Performance Comparison}
\label{sec:exp_compare}

\subsubsection{Competing Methods}

For all competing methods, we train their models with their official implementations using their suggested experimental settings and pick the best checkpoints on the validation set for comparisons.

\vspace{-0.8em}
\paragraph{DeepCSR \cite{cruz2021deepcsr}} is an implicit surface-based that directly predicts implicit surface representations for the coordinates in the MRI images. The surface can be reconstructed using either occupancy field or signed distance function. We reproduce the method in both ways and observe that signed distance function yields better performance. 

\vspace{-0.8em}
\paragraph{Vox2Cortex \cite{bongratz2022vox2cortex}} is a deformation-based model proposed to retrieve cortical surfaces by deforming a generic template and employing a joint graph neural network and U-Net. The results shown below are generated based on the higher-resolution templates with $\approx 168,000$ vertices for each structure. 
Our framework is built upon Vox2Cortex \cite{bongratz2022vox2cortex}, and therefore it serves as our baseline model.

\vspace{-0.8em}
\paragraph{CorticalFlow++ \cite{lebrat2022corticalflow}} is a diffeomorphic-based method that successively deforms template meshes by integrating a discrete stationary velocity grid using traditional ODE solvers. 

\vspace{-0.8em}
\paragraph{CortexODE \cite{ma2022cortexode}} is a multi-stage deformation approach that obtains the white matter initial surface from a volumetric segmentation model, then leverages neural ordinary differential equations to deform an initial WM surface into refined WM and pial surfaces by learning a diffeomorphic flow.

\subsubsection{Evaluation Metrics}
We evaluate our method as well as other approaches using three commonly used metrics including average symmetric surface distance (ASSD), normal consistency (NC), and self-intersection faces ratio (SI). To calculate ASSD and NC, we sample $100$K points uniformly from both predicted and target meshes. For measuring regularity, we determine SI faces using PyMesh \cite{zhou2019pymesh} library. 
\subsubsection{Results Discussion}

\vspace{-0.8em}
\paragraph{Accuracy}
\begin{table*}
\centering
\caption{
\textbf{Cortical Surface Reconstruction Performance Comparison} in terms of average symmetric surface distance (ASSD), normal consistency (NC), and self-intersection face ratio (SI) on ADNI and OASIS datasets. Best values are highlighted. ASSD results are in mm. All results are listed in the format ``mean value $\pm$ standard deviation". \textbf{Hybrid-CSR} represents the cortical reconstruction results $\hat{M}$ without any topology correction operation. \textbf{\qq{+TC}} and \textbf{\qq{+TC+Refine}} indicate the performance of reconstructed cortical surfaces $\hat{M}_{tc}$ and $\hat{M}_{f}$ separately. While $\downarrow$ means smaller metric value is better, $\uparrow$ indicates larger metric value is better.
} 

(a) ANDI dataset

\vspace{0.3em}
\resizebox{\linewidth}{!}{%
\begin{tabular}{lcccccccccccc} 
\toprule
               & \multicolumn{3}{c}{Left Pial} 
               & \multicolumn{3}{c}{Left WM} 
               & \multicolumn{3}{c}{Right Pial} 
               & \multicolumn{3}{c}{Right WM}  \\ 
               
\cmidrule(lr){2-4}\cmidrule(lr){5-7}\cmidrule(lr){8-10}\cmidrule(lr){11-13}
          Method & ASSD (mm) $\downarrow$ & NC $\uparrow$    & SI (\%) $\downarrow$    & ASSD (mm) $\downarrow$ & NC $\uparrow$    & SI (\%) $\downarrow$ & ASSD (mm) $\downarrow$ & NC $\uparrow$   & SI (\%) $\downarrow$ & ASSD (mm) $\downarrow$ & NC $\uparrow$  & SI (\%) $\downarrow$             \\ 
\midrule
DeepCSR~\cite{cruz2021deepcsr}              & .368 $\pm .082$   & .908 $\pm .015$  & \textbf{0}   & .390 $\pm .162$  & .934 $\pm .016$  & \textbf{0}    & .394 $\pm .083$  & .914 $\pm .012$  & \textbf{0}    & .388 $\pm .172$  & .936 $\pm .014$ & \textbf{0}  \\
Vox2Cortex~\cite{bongratz2022vox2cortex}    & .339 $\pm .055$   & .918 $\pm .010$  & .741 $\pm .221$  & .346 $\pm .073$  & .926 $\pm .011$  & .719 $\pm .214$    & .350 $\pm .037$  & .915 $\pm .009$  & 1.025 $\pm .237$ & .335 $\pm .061$  & .927 $\pm .010$ & .745 $\pm .199$         \\
CorticalFlow++~\cite{lebrat2021corticalflow}& .296 $\pm .079$   & .925 $\pm .011$  & .164 $\pm .093$  & .271 $\pm .071$  & .936 $\pm .009$  & .058 $\pm .032$    & .270 $\pm .044$  & .924 $\pm .010$  & .187 $\pm .104$    & .268 $\pm .073$  & .933 $\pm .009$ & .067 $\pm .032$            \\
CortexODE~\cite{ma2022cortexode}            & .258 $\pm .073$   & .929 $\pm .010$  & .112 $\pm .072$  & \textbf{.234} $\pm .064$  & .938 $\pm .010$  & .013 $\pm .011$   & .214 $\pm .035$  & .927 $\pm .009$  & .173 $\pm .091$  & \textbf{.231} $\pm .052$  & .939 $\pm .009$  & .004 $\pm .005$          \\
\cmidrule(lr){1-13}
Hybrid-CSR (ours)                           & .254 $\pm .054$   & .886 $\pm .012$  & \textbf{0}    & .264 $\pm .055$  & .899 $\pm .010$  &  \textbf{0}    & .250 $\pm .041$  & .886 $\pm .012$  &  \textbf{0}    & .257 $\pm .045$  & .901 $\pm .010$ & \textbf{0}              \\
+ TC                                        & .267 $\pm .056$   & .926 $\pm .010$  & \textbf{0}  & .268 $\pm .056$  & \textbf{.939} $\pm .010$  & \textbf{0}    & .262 $\pm .042$  & .926 $\pm .010$  & \textbf{0}     & .260 $\pm .046$  & .940 $\pm .010$ & \textbf{0}           \\
+ TC + Refine                               & \textbf{.203} $\pm .049$   & \textbf{.935} $\pm .009$  & .090 $\pm .075$  & .244 $\pm .056$  & \textbf{.939} $\pm .010$  & .042 $\pm .025$   & \textbf{.200} $\pm .034$  & \textbf{.935} $\pm .008$ & .080 $\pm .071$    & .240 $\pm .043$  & \textbf{.941} $\pm .010$ & .011 $\pm .010$    \\ 
\bottomrule
\end{tabular}
}

\vspace{1em}

(b) OASIS dataset

\vspace{0.3em}

\resizebox{\linewidth}{!}{%
\begin{tabular}{lcccccccccccc} 
\toprule
               & \multicolumn{3}{c}{Left Pial} 
               & \multicolumn{3}{c}{Left WM} 
               & \multicolumn{3}{c}{Right Pial} 
               & \multicolumn{3}{c}{Right WM}  \\ 
               
\cmidrule(lr){2-4}\cmidrule(lr){5-7}\cmidrule(lr){8-10}\cmidrule(lr){11-13}
          Method & ASSD (mm) $\downarrow$ & NC $\uparrow$    & SI (\%) $\downarrow$    & ASSD (mm) $\downarrow$ & NC $\uparrow$    & SI (\%) $\downarrow$ & ASSD (mm) $\downarrow$ & NC $\uparrow$   & SI (\%) $\downarrow$ & ASSD (mm) $\downarrow$ & NC $\uparrow$  & SI (\%) $\downarrow$             \\ 
\midrule
DeepCSR                     & .424 $\pm .075$   & .898 $\pm .016$   & \textbf{0} & .312 $\pm .124$  & .941 $\pm .010$ & \textbf{0} & .444 $\pm .087$  & .895 $\pm .018$ & \textbf{0} & .344 $\pm .158$  & .941 $\pm .011$  & \textbf{0}          \\
Vox2Cortex                  & .401 $\pm .040$   & .900 $\pm .012$   & 1.110 $\pm .270$ & .302 $\pm .037$  & .928 $\pm .008$ & .994 $\pm .193$ & .405 $\pm .044$  & .898 $\pm .012$ & 1.321 $\pm .252$ & .303 $\pm .042$  & .929 $\pm .009$ & 1.022 $\pm .186$           \\
CorticalFlow++ ~~~~~~~~     & .326 $\pm .058$   & .913 $\pm .011$   & .147 $\pm .100$ & .225 $\pm .038$  & .937 $\pm .007$ & .054 $\pm .060$ & .318 $\pm .057$  & .913 $\pm .011$ & .192 $\pm .123$ & .227 $\pm .046$  & .935 $\pm .008$ & .076 $\pm .068$           \\
CortexODE                   & .279 $\pm .052$   & .919 $\pm .009$   & .277 $\pm .096$ & \textbf{.183} $\pm .036$  & \textbf{.943} $\pm .007$   & .032 $\pm .025$ & .280 $\pm .052$   & .918 $\pm .010$    & .151 $\pm .060$ & \textbf{.182} $\pm .052$  & \textbf{.943} $\pm .008$ & .022 $\pm .020$           \\
\cmidrule(lr){1-13}
Hybrid-CSR (ours)           & .298 $\pm .045$   & .879 $\pm .013$   & \textbf{0} & .220 $\pm .039$  & .905 $\pm .008$ & \textbf{0} & .301 $\pm .049$  & .880 $\pm .013$    & \textbf{0} & .218 $\pm .047$  & .907 $\pm .009$ & \textbf{0}             \\
+ TC                        & .311 $\pm .045$   & .915 $\pm .011$   & \textbf{0} & .220 $\pm .039$  & .941 $\pm .008$ & \textbf{0} & .313 $\pm .049$  & .915 $\pm .012$    & \textbf{0} & .219 $\pm .046$  & .942 $\pm .009$ & \textbf{0}           \\
+ TC + Refine               & \textbf{.274} $\pm .043$   & \textbf{.921} $\pm .010$ & .034 $\pm .022$ & .198 $\pm .035$  & \textbf{.943} $\pm .008$  & .040 $\pm .022$ & \textbf{.275} $\pm .052$  & \textbf{.920} $\pm .010$  & .029 $\pm .018$ & .199 $\pm .033$  & \textbf{.943} $\pm .008$  & .037 $\pm .026$   \\ 
\bottomrule
\vspace{-0.8em}
\end{tabular}
}

\label{tab:benchmark}
\end{table*}
As shown in Tab.~\ref{tab:benchmark}, our proposed method surpasses other competing methods in terms of surface reconstruction accuracy (ASSD and NC). 
Compared with any implicit-based and explicit-based methods including DeepCSR, Vox2Cortex, and CorticalFlow++, our hybrid approach without refinement can generate significantly more accurate results on all cortical surfaces.
Furthermore, we got comparable performance with CortexODE on both ADNI and OASIS datasets. While CortexODE performs slightly better than us on in terms of ASSD on WM surfaces, we excel them in terms of ASSD on pial surfaces and NC on all surfaces. We observe that WM surfaces in CortexODE have a higher performance than our Hybrid-CSR because it is based on volumetric segmentation models which are typically suitable for extracting structures with details but few ambiguities in topology, i.e., white matter. Our method, on the other hand, performs better in more diverse shapes. 

\vspace{-0.8em}
\paragraph{Regularity}
To assess the regularity of Hybrid-CSR, we calculate the percentage of self-intersecting faces and present them in Tab.~\ref{tab:benchmark}. 
Thanks to the non-self-intersection property of Marching Cubes, we guarantee to achieve $0$ self-intersection, same as DeepCSR \cite{cruz2021deepcsr}. 
Although a small number of self-intersecting faces are introduced after surface refinement, we still significantly surpass Vox2Cortex, CorticalFlow++, and CortexODE on all surfaces, except for CortexODE \cite{ma2022cortexode} on WM surfaces.

\vspace{-0.8em}
\paragraph{Consistency}
\label{exp:consistency}

\begin{table}
\centering
\caption{
\textbf{Cortical Surface Reconstruction Consistency Comparison} in terms of ASSD on TRT dataset. Hybrid-CSR here indicates the result going through the whole post processing procedures.
}
\resizebox{\columnwidth}{!}{
\begin{tabular}{lccr}
    \toprule
        Method & ASSD (mm) & $>1$mm & $>2$mm  \\
    \midrule
        Vox2Cortex & .301 $\pm .171$ & 3.12\% & .69\% \\
        CortexODE & .280 $\pm .164$ &  2.21\%   &  .38\%\\
        FreeSurfer & .301 $\pm .176$ & 3.35\%   & .94\% \\
    \midrule
        Hybrid-CSR + TC + Refine & \textbf{.258} $\pm .051$  & \textbf{1.61}\%  & \textbf{.18}\%  \\
    \bottomrule
\vspace{-0.8em}
\end{tabular}
}

\label{tab:consistency}
\end{table}
We evaluate the consistency of Hybrid-CSR (with topology correction and surface refinement), Vox2Cortex \cite{bongratz2022vox2cortex}, and CortexODE \cite{ma2022cortexode} (which are all trained on OASIS), and Freesurfer on TRT dataset. 
We generate cortical surfaces from MRI images of the same subject on the same day and measure the ASSD of the resulting reconstructions. The brain morphology of two consecutive scans taken on the same day should be similar to each other, except for the variations caused by the imaging process. 
The result from Table \ref{tab:consistency} shows that we outperform Vox2Cortex \cite{bongratz2022vox2cortex}, CortexODE \cite{ma2022cortexode}, and Freesurfer on the consistency aspect.

\subsection{Results Visualization}
\label{sec:vis}
\begin{figure}
\centering
\includegraphics[width=0.47\textwidth]{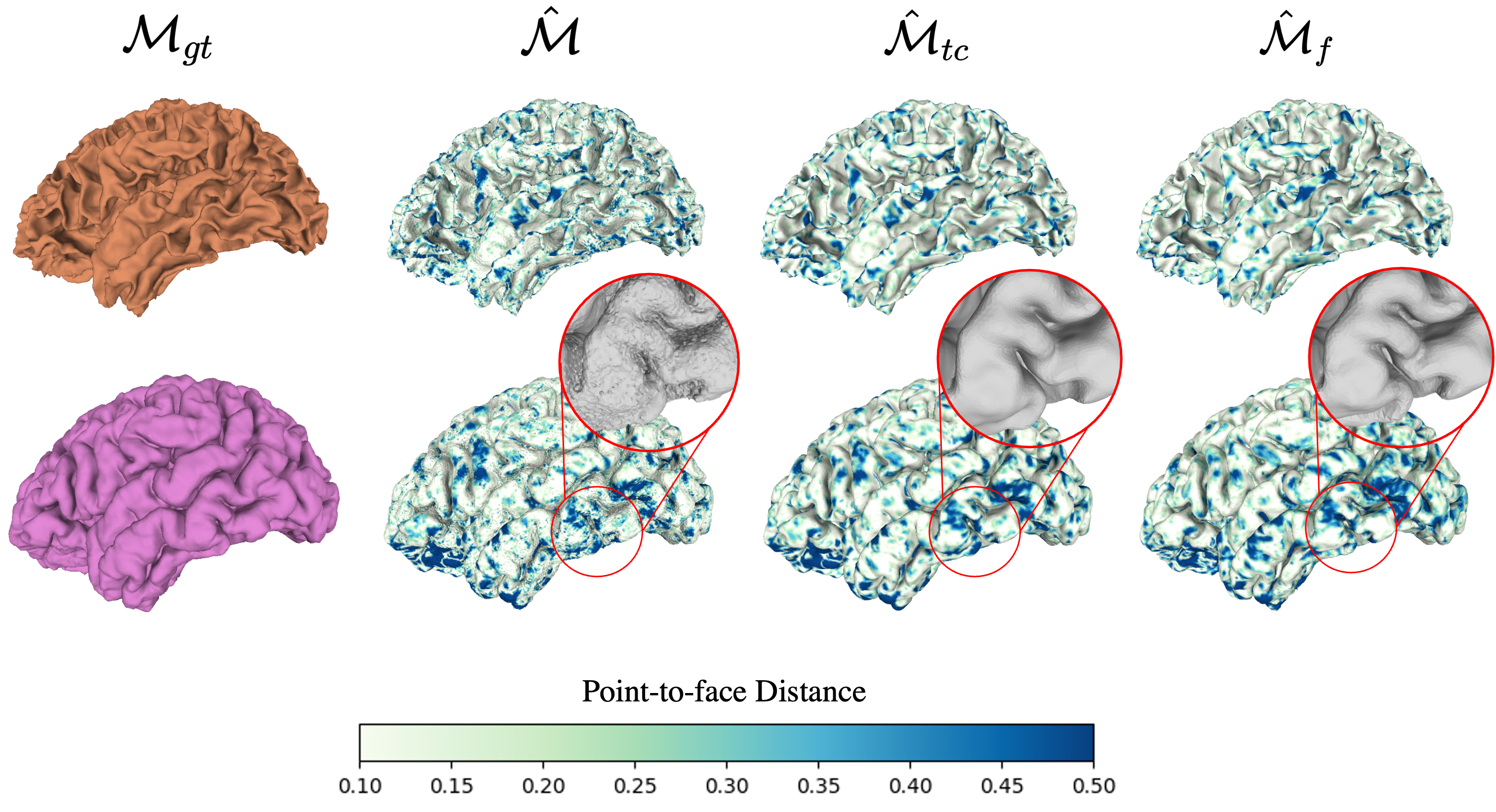}

\caption{\textbf{Hybrid-CSR Results Visualization.} $\mathcal{M}_{gt}$ is the ground truth. $\hat{\mathcal{M}}$ refers to the meshes reconstructed from poisson surface reconstruction. $\hat{\mathcal{M}}_{tc}$ represents the meshes going through topology correction. And $\hat{\mathcal{M}}_f$ indicates the refined meshes. Colors indicate point-to-surface distances, with darker colors indicating larger prediction errors.}

\label{fig:vis}
\end{figure}

The cortical surface reconstruction results generated by Hybrid-CSR are shown in Fig.~\ref{fig:vis}, where $\hat{\mathcal{M}}$ denotes the surfaces reconstructed from PSR that are already well-aligned to the ground truth. After topology correction, $\hat{\mathcal{M}}_{tc}$ is smoother and topologically correct, and the surface refinement procedure further completes the details in the cortical reconstruction, as shown in $\hat{\mathcal{M}}_f$. More visualization of Hybrid-CSR with competing methods would be presented in supplementary material.
\subsection{Ablation Study}
\label{sec:ablation}

We conduct two ablation experiments of WM surface reconstruction on OASIS validation set. For fair comparisons, all experiments below are based on the same coarse mesh deformation module pre-trained for 50 iterations, then the oriented point cloud module will be finetuned for another 50 iterations using different settings. The evaluation metrics are ASSD and 90-percentile Hausdorff distance (HD90).

\vspace{-0.8em}
\begin{table}
\centering
\caption{
\textbf{Ablation Study on Upsample Ratio} $\boldsymbol{S}$. 
}
\resizebox{\columnwidth}{!}{
\begin{tabular}{lcccccc}
\toprule
& & \multicolumn{2}{c}{Left WM} & \multicolumn{2}{c}{Right WM} \\
\cmidrule(lr){3-4}\cmidrule(lr){5-6}
Method  & \#~params   & ASSD (mm)         & HD90 (mm)     & ASSD (mm)      & HD90 (mm)     \\ 
\midrule
    $S=1$              & 6.49M & .357 $\pm .034$  & .881 $\pm .109$ & .352 $\pm .039$ & .867 $\pm .114$  \\
    $S=4$              & 6.54M & .323 $\pm .036$ & .727 $\pm .105$ & .317 $\pm .042$ & .712 $\pm .110$  \\
    $S=7$              & 6.59M & .314 $\pm .036$ & .692 $\pm .107$ & .309 $\pm .044$ & .678 $\pm .116$ \\
    $S=10$             & 6.64M & \textbf{.312} $\pm .037$ & \textbf{.688} $\pm .107$ & \textbf{.308} $\pm .043$ & \textbf{.669} $\pm .113$ \\
\bottomrule
\vspace{-0.8em}
\end{tabular}
}

\label{tab:upsample}
\end{table}
\begin{table}
\centering
\caption{
\textbf{Ablation Study on Network Type} 
}
\resizebox{\columnwidth}{!}{
\begin{tabular}{ccccc}
\toprule
Network & \multicolumn{2}{c}{Left WM} & \multicolumn{2}{c}{Right WM} \\
\cmidrule(lr){2-3}\cmidrule(lr){4-5}
Type     & ASSD          & HD90      & ASSD      & HD90      \\ 
\midrule
    GCN        & .314 $\pm .036$ & .692 $\pm .107$ & .309 $\pm .044$ & .678 $\pm .116$ \\
    GLU        & \textbf{.304} $\pm .036$ & \textbf{.661} $\pm .108$ & \textbf{.298} $\pm .044$ & \textbf{.649} $\pm .114$ \\
\bottomrule
\end{tabular}
}

\label{tab:glu}
\end{table}
\paragraph{Hybrid Representation and Upsample Ratio}  In this study, we employed GCN to estimate point positions and normals for Hybrid-CSR with varying upsample ratios denoted by $S$. As shown in Table~\ref{tab:upsample}, increasing the upsample ratio results in more accurate reconstruction of the white matter surface, while not significantly increasing the number of parameters that need optimization. We have chosen the value of $S=7$ in our experiments to strike a balance between training efficiency and accuracy of the reconstructed surfaces.

\vspace{-0.8em}
\paragraph{GLU Estimation} In this study, we apply two different networks, i.e., GCN and GLU, to estimate the oriented point clouds. As can be observed from Tab~\ref{tab:glu}, GLU can generate significantly more accurate reconstruction results on WM surfaces. 
GCN tends to generate over-smoothing predictions, while GLU can produce more acute deformations. Since PSR is robust to outlier points, GLU might be an advantageous choice. More discussions on GLU-based point cloud estimation are included in the supplementary.

\section{Limitations and Further Directions}

Our proposed method for cortical surface reconstruction has some limitations that need to be considered. One issue is the running time, as our method uses neural fields-based diffeomorphic surface registration to correct topology, which takes about 2 minutes for 75-iteration optimizations. This makes it slower than methods \cite{lebrat2021corticalflow, santa2022corticalflow++, bongratz2022vox2cortex, hoopes2021topofit} relying on explicit surface reconstruction. To address this issue, a persistent homology prior can be imposed on the reconstructed indicator field to further eliminate the need for topology correction procedures in our proposed method. Additionally, memory consumption can be reduced by handling anatomical structures separately or sequentially. CorticalFlow++ has shown that reconstructing the gray matter surface starting from the white matter surface can be beneficial. These improvements will help us achieve even more accurate and efficient cortical surface reconstruction.

Another limitation lies in the data sources, where we used pseudo labels generated from Freesurfer without manually removing faulty labels from our training and test set. This may lead to inaccurate evaluation results and potentially overfitting to some noisy labels. Additionally, our method is not designed to handle MRI scans with significant domain shifts. That is, Hybrid-CSR may generate undesired results when applied to brain MRI scans with tumors or other morphological changes.

\section{Conclusion}
\label{sec:conclusion}

This paper introduces Hybrid-CSR, a novel cortical surface reconstruction framework that leverages both explicit and implicit shape representations through differentiable Poisson surface reconstruction. The proposed method includes a new topology correction pipeline that employs continuous diffeomorphic flow optimization. The effectiveness of Hybrid-CSR is evaluated on multiple brain datasets, and it shows competitive performance in terms of accuracy, regularity and consistency. Moreover, the versatility of our method extends beyond the cortical surface reconstruction problem and has the potential for wide application in a variety of surface reconstruction scenarios.

{\small
\bibliographystyle{ieee_fullname}
\bibliography{egbib}
}

\appendix

\setcounter{page}{1}

\twocolumn[
\centering
\Large
\textbf{Supplementary Material} \\
\vspace{1.0em}
] 

\appendix

\begin{figure*}[t]
\centering
\includegraphics[width=\textwidth]{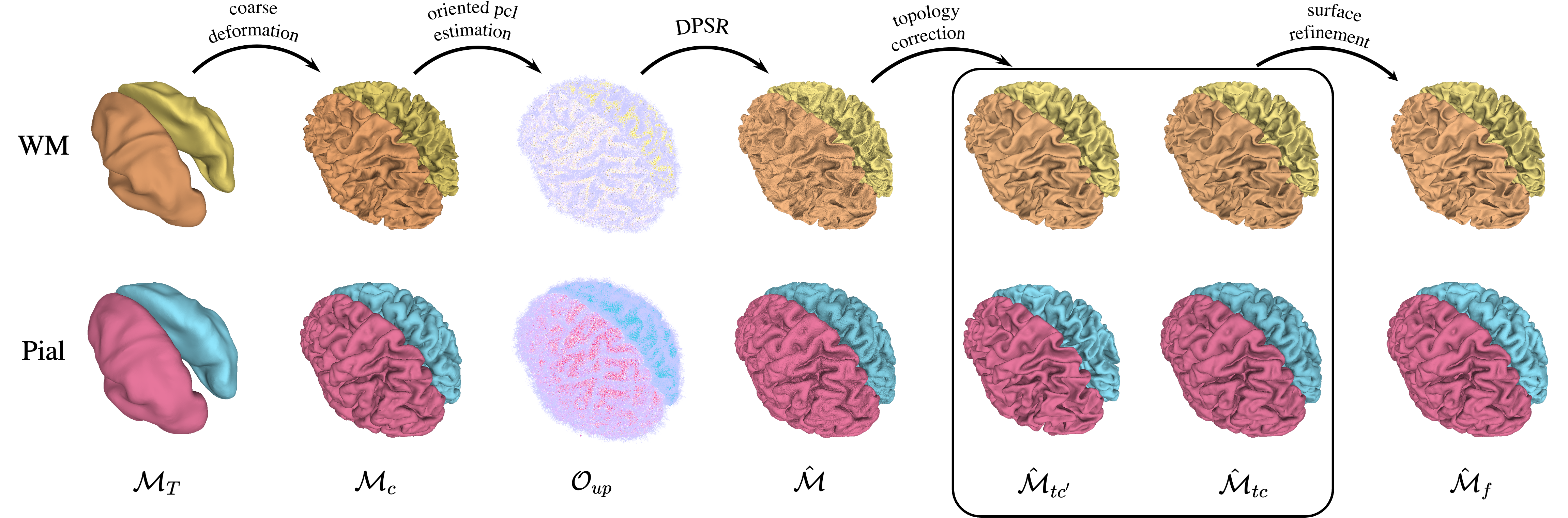}

\caption{\textbf{The pipeline of Hybrid-CSR}.}

\label{fig:pipeline_supp}
\end{figure*}
\section{Pipeline Review}
\label{sec:details}
Fig.~\ref{fig:pipeline_supp} depicts the whole pipeline of Hybrid-CSR. From the template meshes $\mathcal{M}_T$, Hybrid-CSR first obtains coarsely deformed cortical meshes $\mathcal{M}_c$, given which, we estimate the positions and normals of upsampled oriented point cloud $\mathcal{O}_{up}$. Then the cortical surfaces $\hat{\mathcal{M}}$ can be reconstructed via poisson surface reconstruction from $\mathcal{O}_{up}$. To fix the topology defects in $\hat{\mathcal{M}}$, we extract non-zero level set from the signed distance grids, obtaining topologically correct meshes $\hat{\mathcal{M}}_{tc'}$, and apply optimization-based diffeomorphic registration to recover the accurate and smooth genus-0 cortical surfaces $\hat{\mathcal{M}}_{tc}$. Lastly, we refine $\hat{\mathcal{M}}_{tc}$ using a learning-based diffeomorphic transformation model, to achieve our final reconstruction results $\hat{\mathcal{M}}_f$.

\section{Implementation Details}
\label{sec:details}

Hybrid-CSR framework is implemented using PyTorch \cite{paszke2019pytorch} and executed on a system equipped with an NVIDIA RTX A6000 GPU and an Intel i7-7700K CPU.

\subsection{Toy Example}
\label{sec:details_toy}

The target contour is controlled by 40 pivot points and the source circle contour includes 200 pivot points. The mesh deformation is modeled via neural fields \cite{xie2022neural}. The positions and normals of oriented point cloud are optimized directly.

\subsubsection{Network Architecture}
\label{sec:toy_net}

To model contour displacements, we encoder positions with random Fourier mapping \cite{tancik2020fourier} and approximate the deformation field with a SIREN \cite{sitzmann2020implicit} model. The gaussian scale and embedding length of position encoding are 5 and 128. The hidden features size and hidden layers number are 256 and 2. Our hybrid method directly optimizes the positions and normals of oriented points. 

\subsubsection{Optimization}

To optimize neural fields for contour deformation, 1000 points with normals are separately sampled from ground truth contour and deformed coutour in each iteration. The loss function consists of geometry-consistency loss and regularization loss. For the geometry-consistency loss, we add up the chamfer distance $\mathcal{L}_{cd}$ and normal distance $\mathcal{L}_{nd}$ between two sets of 1000 sampled oriented points. The regularization loss consists of edge length $\mathcal{L}_{edge}$ \cite{wang2018pixel2mesh}, as well as normal consistency regularization $\mathcal{L}_{nc}$. The total mesh loss is as below:
\begin{equation}
    \mathcal{L} = \mathcal{L}_{cd} + 0.02*\mathcal{L}_{nd} + 0.005*\mathcal{L}_{edge} + 0.005*\mathcal{L}_{nc}
\end{equation}
We apply Adam optimizer \cite{kingma2014adam} with a learning rate of $1e^{-4}$ for 3000 iterations to update the parameters of neural fields.

For the hybrid method, we first uniformly sample 1000 points with normals from the deformed contour obtained above, as initializations. To optimize the positions and normals of oriented points, we minimize the L2 loss between the indicator map reconstructed from the ground truth and the optimized oriented points. We apply Adam optimizer with a learning rate of $3e^{-3}$ for 1000 iterations.

\subsubsection{Results of diffeomorphic transformation}
\label{sec:diff_opt}

\begin{figure}[H]
\centering
\includegraphics[width=0.47\textwidth]{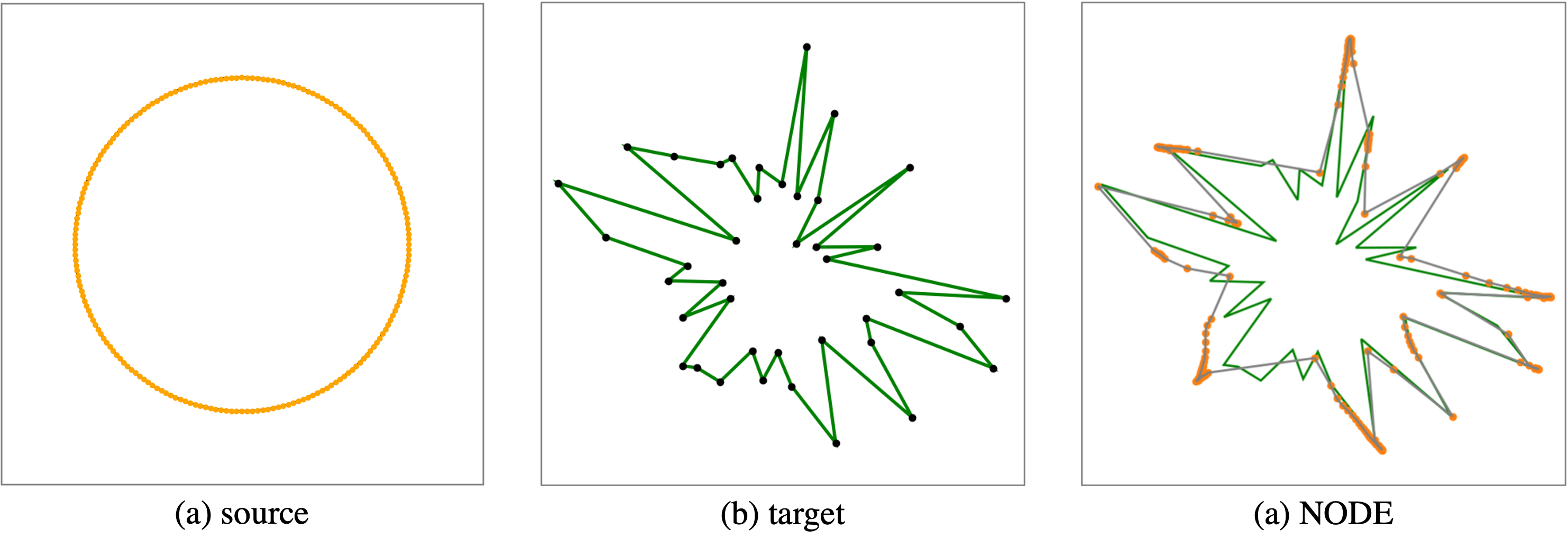}

\caption{Explicit contour representation by diffeomorphic transformtion from circle source contour. 
}

\label{fig:toy_example_supp}
\end{figure}

In NMF \cite{gupta2020neural}, they present that diffeomorphic transformation can avoid the ``regularizer`s dilemma'', but in our toy experiment, we found neural ode (NODE) \cite{chen2018neural} is not suitable to model large and sharp deformations, as is shown in Fig.~\ref{fig:toy_example_supp}. We model the dynamic function of NODE, i.e., neural velocity fields, using the same neural fields as that for contour deformation and optimize with only chamfer distance and normal distance without any regularization loss, optimized via Adam optimizer with a learning rate of $1e^{-4}$ for 3000 iterations.

\subsection{Coarse Mesh Deformation}
\label{sec:details_cmd}

\subsubsection{Network Architecture}

We apply Vox2Cortex \cite{bongratz2022vox2cortex} to deform template meshes. Same as Vox2Cortex, 
\begin{itemize}
    \item we train the volumetric segmentation branch and mesh deformation branch end-to-end;
    \item volumetric segmentation branch is based on the Res-Unet;
    \item we take four cortical surfaces template as one and use residual GCN-based modules to encode features on graphs;
    \item we use the same feature extraction strategy. That is, in the first step of mesh deformation, the volumetric feature associated with meshes vertices are from the 4th, 5th, 6th and 7th layers of Res-Unet. And in the second step of mesh deformation, the meshes vertices are extracted from the 3rd, 4th, 7th and 8th layers of Res-Unet;
    
\end{itemize}

Different from Vox2Cortex, in the coarse mesh deformation module of Hybrid-CSR 
\begin{itemize}
    \item we deform template meshes in two steps, instead of four steps. In other words, our superiority in performance doesn't come from more steps of surface reconstruction;
    \item In both training and inference, we use smaller templates ($\approx 42000$ vertices per surface).
    
\end{itemize}

\subsubsection{Training}

Same as Vox2Cortex, we apply the loss function composed of voxel loss $\mathcal{L}_vox$, curvature-weighted chamfer loss $\mathcal{L}$, normal distance, laplacian smoothing, normal consistency as well as edge length regularizations. The coarse mesh deformation module as well as segmentation branch are first optimized for 50 epochs and then they will be optimized together with the oriented point cloud estimation module for another 100 epochs. The other implementation details can be found in the supplementary material of Vox2Cortex.

\subsection{Oriented Point Cloud Estimation}
\label{sec:details_opcle}

\subsubsection{Gated Linear Unit (GLU) for Point Estimation} 

Let $S$ denote the upsample ratio, $\boldsymbol{p}_i^{up} \in \mathbb{R}^{S \times 3}$ denote the position of upsampled oriented point clouds, $\boldsymbol{v}_i^{up} \in \mathbb{R}^{S \times 3}$ denote the vertex of deformed meshes repeated by $S$ times and $\boldsymbol{p}_i^{up}$ denote the upsampled displacements of vertex $\boldsymbol{v}_i$. 
Let's also define the intermediate position of upsampled oriented point clouds as $\boldsymbol{p'}_i^{up} \in \mathbb{R}^{S \times 3}$, such that 
\begin{equation}
    \boldsymbol{p}_i^{up} = (1-\boldsymbol{m}_i) \odot \boldsymbol{v}_i^{up} + \boldsymbol{m}_i \odot \boldsymbol{p'}_i^{up}
    \label{eq:point_displace_supp_1}
\end{equation}
where each dimension of $\boldsymbol{m}_i$ is between 0 and 1, controlling the ``confidence'' of the intermediate results. 
We can represent $\boldsymbol{p'}_i^{up}$ as $\boldsymbol{v}_i^{up} + \boldsymbol{d'}_i$, where $\boldsymbol{d'}_i^{up}$ is intermediate displacement associated with upsampled vertex $\boldsymbol{v}_i^{up}$. Therefore, the Eq.~\ref{eq:point_displace_supp_1} can be rewritten as:
\begin{equation}
    \boldsymbol{p}_i^{up} = \boldsymbol{v}_i^{up} + \boldsymbol{m}_i \odot \boldsymbol{d'}_i^{up}
    \label{eq:point_displace_supp_2}
\end{equation}
As we have demonstrated in the main paper, using GLU, the displacements can be written as $\boldsymbol{d_i} = (\mathbf{W}_0 \mathbf{f}_i+\mathbf{b}_0) \odot \sigma(\mathbf{W}_1 \mathbf{f}_i+\mathbf{b}_1)$. Then, we have
\begin{equation}
    (\mathbf{W}_0 \mathbf{f}_i+\mathbf{b}_0) \odot \sigma(\mathbf{W}_1 \mathbf{f}_i+\mathbf{b}_1) = \boldsymbol{m}_i \odot \boldsymbol{d'}_i^{up}
    \label{eq:point_displace_supp_2}
\end{equation}
Thus, we suppose $\mathbf{W}_0 \in \mathbb{R}^{(S \times 3) \times (d_{out} + S \times 3)}$ and $\mathbf{b}_0 \in \mathbb{R}^{(S \times 3)}$ modulate the displacement vector, while $\mathbf{W}_1 \in \mathbb{R}^{(S \times 3) \times (d_{out} + S \times 3)}$ and $\mathbf{b}_1 \in \mathbb{R}^{(S \times 3)}$ modulate the ``confidence '' of the predicted displacement vector. In our experiments, $S=7$ and $d_{out}=64$. 

\subsubsection{Network Architecture for Normal Estimation} 
\begin{figure*}[!ht]
\centering
\includegraphics[width=0.9\textwidth]{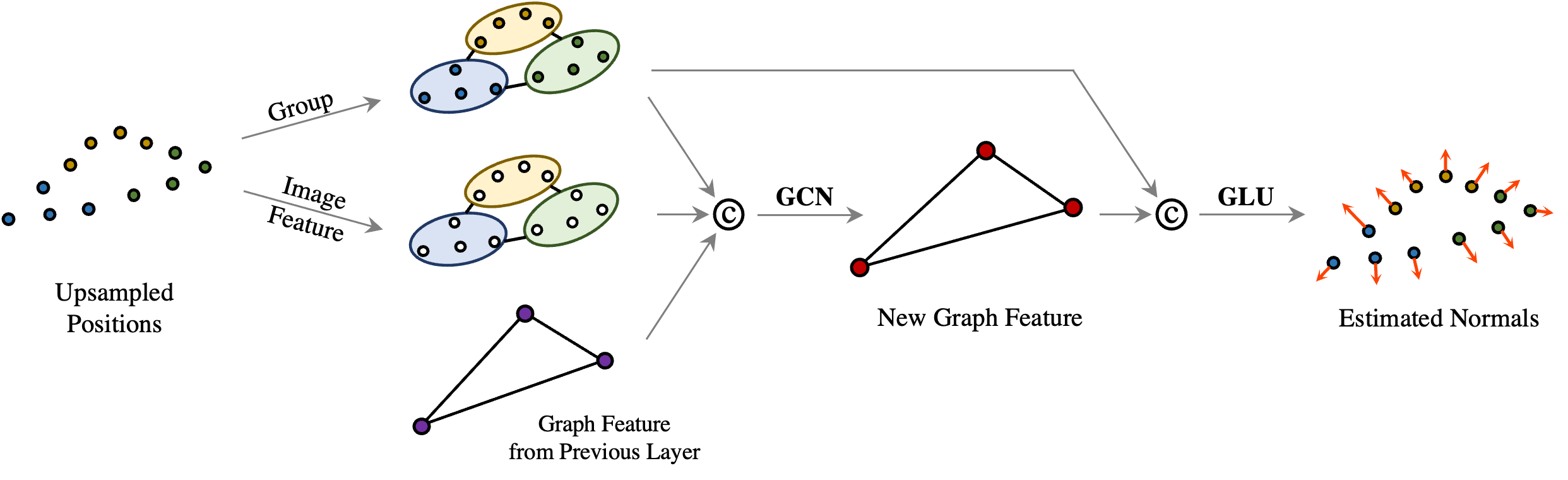}

\caption{\textbf{Network Architecture of Normal Estimation Module.} 
We apply GCN to encode features on graph and GLU to estimate normals of upsampled positions.
} 


\label{fig:normal_est_supp}
\end{figure*}

As is shown in Fig.~\ref{fig:normal_est_supp}, GCN is used to encode features on graph and GLU is used to estimate normals of upsampled positions. The input of GCN is the concatenation of grouped positions, grouped image features associated with points and graph feature generated by the previous GCN layer. The positions and associated features will be grouped into the same node if they are displaced from the same vertices. We sample the image features from the 2nd, 3rd, 8th and 9th layers of Res-Unet given the point positions via linear interpolation, and the channel number of these volumetric features are 32, 64, 16 and 8. The graph feature from the previous layer is in length of 64. Thus, the input feature channel number of GCN is $S*(3+(32+64+16+8)) + 64 = 925$ and the output feature channel number is 64. 

The output of GCN concatenated with the grouped positions is taken as the input of GLU, whose input feature channel number is therefore $64+S*3=85$. The output of GLU is the normals of upsampled point clouds, thus the output channel number is $S*3=21$.

\subsubsection{Training}

We apply weighted mean square error $\mathcal{L}_{DPSR}$ to measure the difference between predicted and ground truth indicator grids. The weight map is the smoothed edge map of the ground truth indicator grid. Together with mesh-based loss proposed in Vox2Cortex, $\mathcal{L}_{DPSR}$ is used to optimize Hybrid-CSR in an end-to-end manner for additional 100 epochs. The parameters of oriented point cloud estimation module are optimized via an Adam optimizer of a learning rate of $5e^{-5}$.

\subsection{Topology Correction}
\label{sec:details_topo}

We model the neural velocity fields $\mathcal{F}_{\theta}$, using the same network architecture as described in Sec.~\ref{sec:toy_net} and Sec.~\ref{sec:diff_opt}. Other implementation details are included in the main paper. 

Different from the toy example, in the procedure of topology correction, the source surface $\hat{\mathcal{M}}_{tc'}$ and target surface $\hat{\mathcal{M}}_{tc}$ have already been well-aligned, so that diffeomorphic transformation optimized by chamfer distance is able to provide accurate surface registration performance.

\subsection{Surface Refinement}
\label{sec:details_refine}

The network architecture of surface refinement is the same as CortexODE \cite{ma2022cortexode}. But there are some differences in training pial surface refinement model. In the original CortexODE, they learn to map the ground truth WM surface to the ground truth pial surface. Since the ground truth WM and pial surface share the same topology, they can train with the L2 distance. However, during inference, the source surface is the predicted WM surface obtained from Marching Cubes so there exists a discrepancy between inference and training. Instead, we learn to map the topological correct pial surface to the ground truth pial surface, supervised by chamfer distance. And during inference, the initial pial surface is also generated by the topology correction procedure. In terms of other implementation details, we follow the original CortexODE.




\section{Visual Comparisons with Competing Methods}
\label{sec:vis_comp}
\begin{figure*}
\centering
\includegraphics[width=\textwidth]{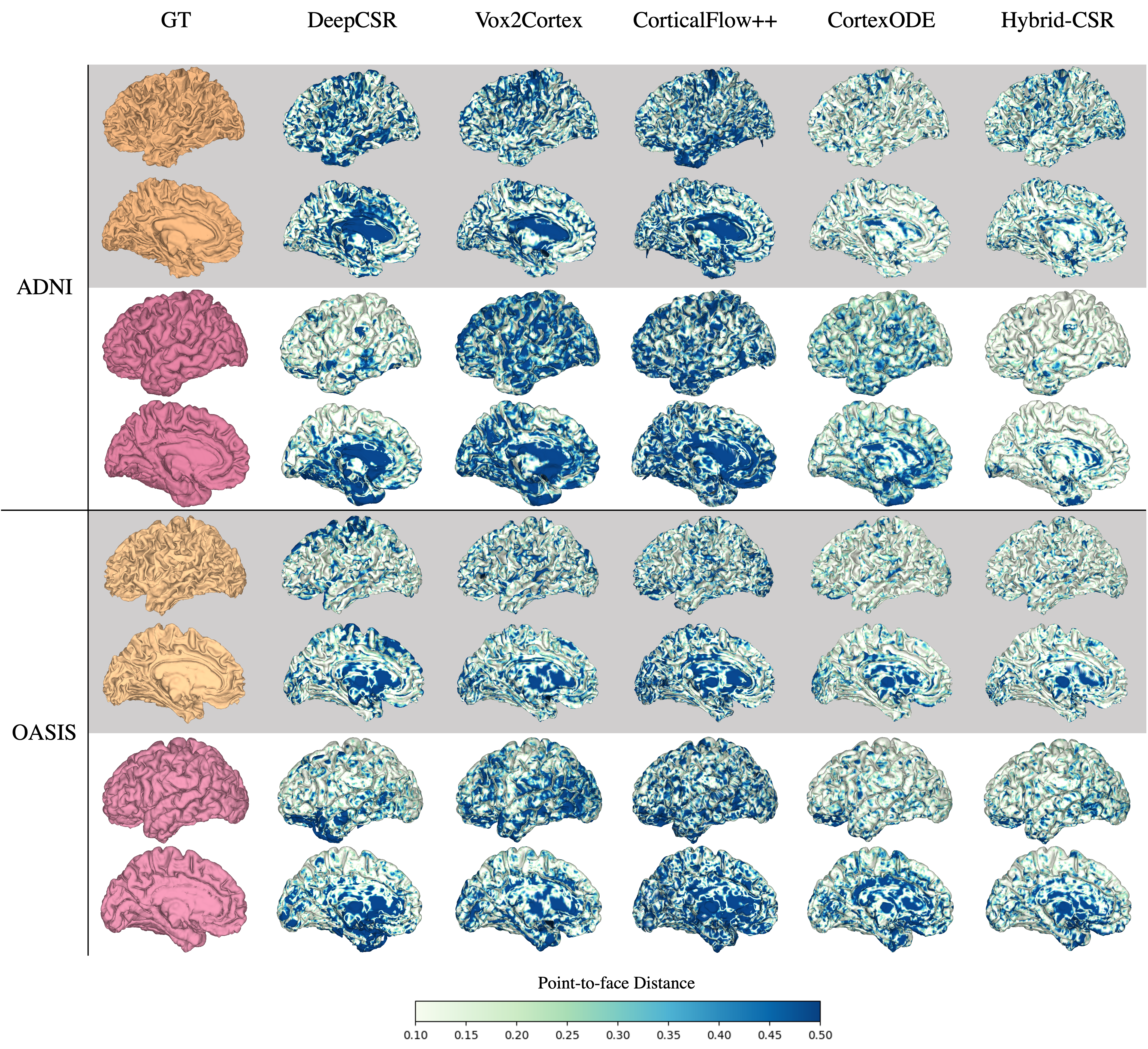}

\caption{\textbf{Visual Comparison between Hybrid-CSR and Other Competing Methods}}

\label{fig:vis_comp}
\end{figure*}

In this section, we provide the comprehensive visual comparison between our Hybrid-CSR and other competing methods, including Vox2Cortex \cite{bongratz2022vox2cortex}, CorticalFlow++ \cite{santa2022corticalflow++}, CortexODE \cite{ma2022cortexode} as well as DeepCSR \cite{cruz2021deepcsr}. From Fig.~\ref{fig:vis_comp}, we can see our proposed Hybrid-CSR in general can generate surfaces in lighter colors, i.e., smaller point-to-surface distances, compared with other methods.

\end{document}